\documentclass{article}

\usepackage{microtype}
\usepackage{graphicx}
\usepackage{subfigure}
\usepackage{booktabs} 
\usepackage{dirtytalk}

\usepackage{hyperref}

\usepackage[accepted]{icml2019}

\icmltitlerunning{Evaluating the distribution learning capabilities of GANs}

\begin{document}

\twocolumn[
\icmltitle{Evaluating the distribution learning capabilities of GANs}



\icmlsetsymbol{equal}{*}

\begin{icmlauthorlist}
\icmlauthor{Amit Rege}{cub}
\icmlauthor{Claire Monteleoni}{cub}
\end{icmlauthorlist}

\icmlaffiliation{cub}{Department of Computer Science, University of Colorado Boulder, Boulder, USA}

\icmlcorrespondingauthor{Amit Rege}{amit.rege@colorado.edu}

\icmlkeywords{GAN, evaluation}

\vskip 0.3in
]



\printAffiliationsAndNotice{}  

\begin{abstract}
We evaluate the distribution learning capabilities of generative adversarial networks by testing them on synthetic datasets. The datasets include common distributions of points in $R^n$ space and images containing polygons of various shapes and sizes. We find that by and large GANs fail to faithfully recreate point datasets which contain discontinous support or sharp bends with noise. Additionally, on image datasets, we find that GANs do not seem to learn to count the number of objects of the same kind in an image. We also highlight the apparent tension between generalization and learning in GANs.
\end{abstract}

\section{Introduction}

Generative Adversarial Models (GANs)\cite{Goodfellow:2014:GAN:2969033.2969125} have been found to produce images of very high quality on some datasets \cite{DBLP:conf/iclr/KarrasALL18, DBLP:journals/corr/abs-1812-04948}. However, their results on other datasets, while impressive, still lag behind \cite{DBLP:journals/corr/abs-1809-11096}. This raises the question whether GANs are indeed the right choice to model some distributions. This paper aims to test the distribution learning ability of GANs by evaluating them on synthetic datasets.

\subsection{Related works and Contributions}
It has been proposed in recent work that \say{a high number of classes is what makes ImageNet \cite{imagenet_cvpr09} synthesis difficult for GANs} \cite{pmlr-v70-odena17a}. Indeed, GANs have been able to produce very high quality images on CelebA \cite{DBLP:journals/corr/abs-1812-04948, DBLP:conf/iclr/KarrasALL18} while results on Imagenet are not so impressive \cite{DBLP:journals/corr/abs-1809-11096}. Because distributions of natural images are complex, in this work, we focus our attention on synthetically generated datasets. We study the learnability of commonly encountered distributions in low dimensional space and the impact of discontinuity. Additionally, we evaluate a specific aspect of learning high dimensional image distributions, counting similar objects in a scene. This constitutes an important part of learning latent space representations of images since for an image to be semantically well-formed, certain objects must obey certain numerical constraints (for example, number of heads on an animal).

Our evaluation is performed on synthetic point and image datasets. To our knowledge, the only instance of synthetic image datasets used for GAN evaluation have been to learn manifolds of convex polygons (specifically triangles) \cite{Lucic:2018:GCE:3326943.3327008}. Although, we also use polygons as a testbed for our experiments, we focus on learning a manifold with multiple polygons where their number is fixed.

Our contributions are as follows:
\begin{enumerate}
    \item We show via experiments on synthetic datasets that commonly found distributions are learnable by GANs. We also highlight that distributions with gaps in support may be difficult to learn without using a mixture of generators.
    \item We empirically evaluate whether GANs can learn semantic constraints on objects in images in high dimensional space. Specifically, we test a GAN's ability to count an object that is repeated in an image.
    \item We underline a possible tension between generalization ability of GANs and their learning capabilities.
\end{enumerate}

\section{Experimental Setup}
In this section, we describe the setup of our experiments which include details about the datasets generated, architectures used and the reasoning behind them. 

\subsection{Datasets}
We generate two kinds of datasets (each with 5000 examples) for our evaluation : point datasets, where each sample is a point in $R^n$ and image datasets with each image containing a fixed number of polygons. We use 4 point datasets in our evaluation : Mixtures of Gaussians, Concentric Circles, S-shape curves and Swiss rolls. The first two are 2D while the latter two are in 3D space. This choice was made to enable us to visualize the learned distribution. For each of these four settings, we experimented with three variants, each containing a different amount of noise.

\begin{figure}
\centering
\begin{minipage}{.45\linewidth}
  \includegraphics[width=\linewidth]{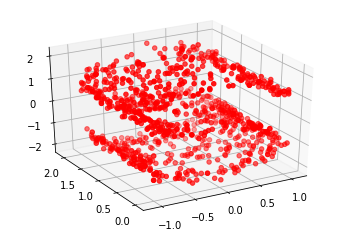}
  \caption{ S-Curve Distribution}
  \label{img4}
\end{minipage}
\hspace{.05\linewidth}
\begin{minipage}{.45\linewidth}
  \includegraphics[width=\linewidth]{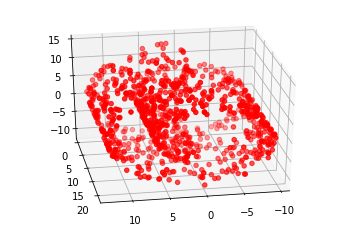}
  \caption{Swiss Roll Distribution}
  \label{img5}
\end{minipage}
\end{figure}

\begin{figure}
\centering
\begin{minipage}{.25\linewidth}
  \includegraphics[width=\linewidth]{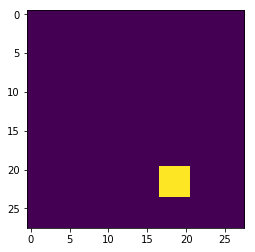}
  \caption{Image from the Squares 1-4 dataset}
  \label{img1}
\end{minipage}
\hspace{.05\linewidth}
\begin{minipage}{.25\linewidth}
  \includegraphics[width=\linewidth]{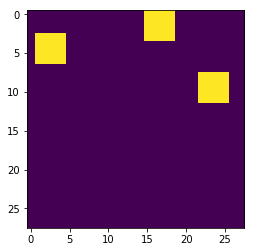}
  \caption{Image from the Squares 3-4 dataset}
  \label{img2}
\end{minipage}
\hspace{.05\linewidth}
\begin{minipage}{.25\linewidth}
  \includegraphics[width=\linewidth]{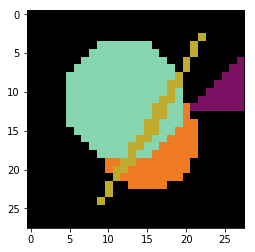}
  \caption{Image from the CT2 dataset}
  \label{img3}
\end{minipage}
\end{figure}

To evaluate high dimensional learning, we generate image datasets which are mixtures of polygons. We created three datasets with each image containing: 1 square of size 4x4 (called Squares 1-4), 3 squares of size 4x4 (called Squares 3-4) and a mixture of two triangles and two circles (called CT2). All the datasets contain images of size 28x28 with the third one containing 3 channels. Additionally, for the first two datasets, all squares are non-overlapping and have edges which are axis-aligned. CT2, on the other hand, contains overlapping polygons. In each dataset, the number of objects is fixed and the only varying quantity is their position (which varies with a Gaussian distribution). For the square datasets, even the rotation and shape of the objects is held constant. Hence, the only source of variation is their position. Some examples are shown in Figures \ref{img1}, \ref{img2}, \ref{img3}.

The main objective behind creating images with a fixed number of objects was to test whether GANs can learn to count the number of objects in a scene. More specifically, since GANs have shown impressive image generation capabilities on centered image datasets\cite{DBLP:conf/iclr/KarrasALL18}, we want to measure whether that performance can transfer to datasets with objects occuring at varying locations in the scene. Since the only varying quantity in the square datasets is the position, we would expect GANs with true distribution learning abilities to be able to produce images with the exact numbers of squares at different positions in the image. 

Additional details about our data generation process can be found in the Appendix. 

\subsection{Architectures}
We use two sets of architectures to train our models. For point datasets, we use a Vanilla GAN with a 3 layer MLP for both the generator and discriminator and another model with the same architecture with Wasserstein loss (enforced via gradient penalty) \cite{Gulrajani:2017:ITW:3295222.3295327}. 

For image datasets, we use one model with a DCGAN-inspired \cite{DBLP:journals/corr/RadfordMC15} architecture and another model with the same architecture with Wasserstein loss (enforced via gradient penalty). We do not evaluate Vanilla GANs on our image datasets because they do not seem to be competitive with the other models in our experiments. Further architectural details (including choice of hyperparameters) are described in the Appendix.

\subsection{Experimental Details}
We used the Google Colaboratory environment (12GB RAM Nvidia Tesla K80) for all our experiments in this paper. All models with Wasserstein loss are trained with RMSProp \cite{Tieleman2012} with a learning rate of 0.00005. All others are trained with the ADAM \cite{DBLP:journals/corr/KingmaB14} optimizer with a learning rate of 0.0002. We train models for up to 150k training steps and stop earlier if the model reaches convergence earlier. Increasing training steps beyond 150k were not found to significantly improve sample quality. For each generator update, we update the discriminator 5 times for WGAN-GP inspired architectures. Additional details about our experiments and architectures can be found in the Appendix.

\section{Results and Analysis}
\subsection{Point Data}

\begin{figure}
\centering
\begin{minipage}{.4\linewidth}
  \includegraphics[width=\linewidth]{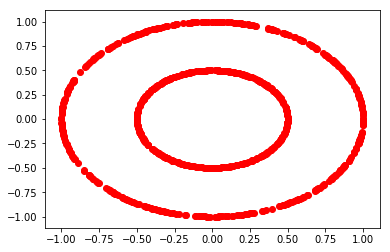}
  \caption{Original distribution of Concentric circles}
  \label{img6}
\end{minipage}
\hspace{.05\linewidth}
\begin{minipage}{.4\linewidth}
  \includegraphics[width=\linewidth]{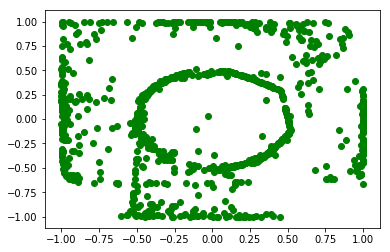}
  \caption{Learned Distribution after 150k steps. \textbf{Note:} this distribution gets better after more iterations but we show the one after 150k for homogeneity}
  \label{img7}
\end{minipage}
\end{figure}

\begin{figure}
\centering
\begin{minipage}{.4\linewidth}
  \includegraphics[width=\linewidth]{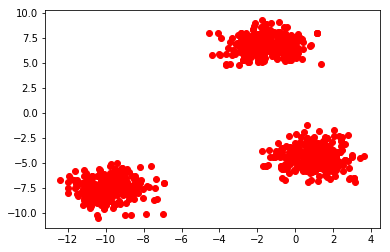}
  \caption{Original distribution of 3 blobs}
  \label{img8}
\end{minipage}
\hspace{.05\linewidth}
\begin{minipage}{.4\linewidth}
  \includegraphics[width=\linewidth]{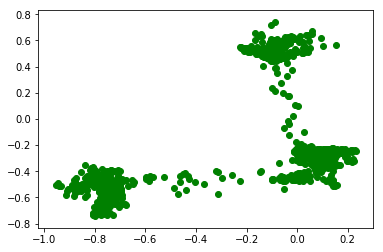}
  \caption{Learned Distribution after 150k steps}
  \label{img9}
\end{minipage}
\end{figure}

On mixtures of Gaussians and concentric circles, we find that both architectures seem to perform equally well. They seem to approach approximate convergence as both their discriminator's accuracy oscillates around 50\%. Since both datasets contain disconnected components, we find that both models are not able to model this discontinuity and as a result still produce samples which lie in between clusters of data. This may explain the oscillatory behaviour observed. Examples of 1000 samples from the real and fake distributions are shown in Figures \ref{img7}, \ref{img9}.

\begin{figure}
\centering
\begin{minipage}{.4\linewidth}
  \includegraphics[width=\linewidth]{Unknown-4.png}
  \caption{Original distribution of the shape S (minimal noise)}
  \label{img10}
\end{minipage}
\hspace{.05\linewidth}
\begin{minipage}{.4\linewidth}
  \includegraphics[width=\linewidth]{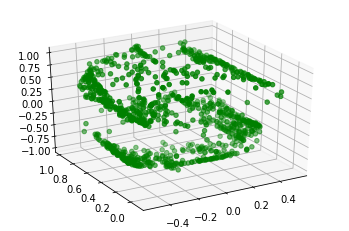}
  \caption{Learned Distribution after 140k steps (minimal noise)}
  \label{img11}
\end{minipage}
\end{figure}

\begin{figure}
\centering
\begin{minipage}{.4\linewidth}
  \includegraphics[width=\linewidth]{Unknown-5.png}
  \caption{Original distribution of the Swiss roll (minimal noise)}
  \label{img12}
\end{minipage}
\hspace{.05\linewidth}
\begin{minipage}{.4\linewidth}
  \includegraphics[width=\linewidth]{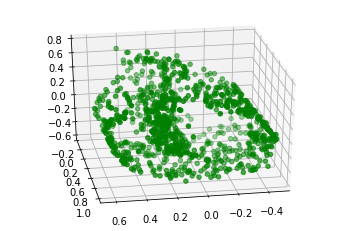}
  \caption{Learned Distribution after 150k steps (minimal noise)}
  \label{img13}
\end{minipage}
\end{figure}

The inability to model discontinuity is understandable since the latent space is continuous and the neural network can be considered to be a continuous function approximator so the output has to be continuous as well. This virtually guarantees that some samples from the model will be necessarily "bad". 

Next, we evaluate the S-curve and Swiss roll distributions. Traditionally, mixtures of Gaussians have been the toy distribution of choice for GAN evaluation. However, we find that both distributions are learned fairly faithfully but increasing noise can cause separate surfaces in the distribution to be merged. For example, with increased noise, the S shape in the S-curve can become an 8 or the Swiss roll may look like a circle (samples in Appendix). 

This phenomenon, in our experiments, seems to be worse (in terms of losing shape) for Swiss rolls than S-curves. We hypothesize this may be due to overlapping surfaces being closer in the Swiss roll distribution making it more sensitive to noise. This dependence of sensitivity to noise on the underlying distribution suggests that GANs may not be suitable to modelling certain distributions in noisy environments and alternative generative models may need to be explored.  

\subsection{Image Datasets}

During training, the first model (without Wasserstein loss) seems to converge after about 74k steps while the other one doesn't seem to converge. Inspite of this difference, we do not find a major difference in image quality from the two models after 150k steps.

\begin{figure}
    \centering
    \includegraphics[width=\linewidth]{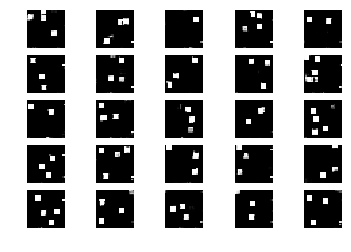}
    \vspace*{-1.2cm}
    \caption{Samples after 150k steps from DCGAN (converged)}
    \label{img14}
    
    \vspace*{0.5cm}
    
    \centering
    \includegraphics[width=\linewidth]{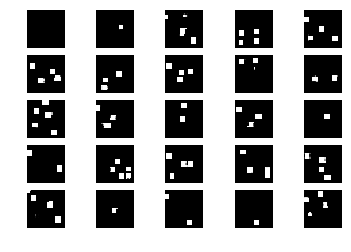}
    \vspace*{-1.2cm}
    \caption{Samples after 150k steps from WGAN-GP}
    \label{img15}
\end{figure}

We find that both models fail to learn to count on the Squares 3-4 dataset. Our models can potentially produce anything between 0 and 5 squares. Some random samples can be seen in \ref{img14}, \ref{img15}. 

This raises an important question about the learning capability of GANs. Is what we are seeing just poor learning or should it be instead viewed as generalization? For example, in natural image datasets, say faces, we might see GANs produce images with completely new hair placement. When these images look "natural" to the observer, they might see it as generalization. However, when that image does not look plausible, we would call it poor learning. As it stands, there seems to be no clear demarcation between poor learning outcomes and generalization.

Since the only source of variation in our image datasets is location, we would like to see the GAN learn that there are 3 squares in each image and then generalize their position. However, currently, we have no way of enforcing this constraint on the number or shape of objects. Quantifying this apparent tradeoff between learning and generalization is an interesting avenue for further work. In fact, there might not even be tradeoff 
and GANs may fundamentally be unable to learn distributions of such a nature. We leave the evaluation of these hypotheses as future work.

We do not focus our discussion excessively on shape because for more natural datasets, shape is more variable and is not rigid, like in our case. For example, we know that cats have 4 legs, but, the legs may not look the same in every image. A natural objection would be that a similar reasoning would be applicable to the count of objects i.e. some of these legs may be occluded in the image and, therefore, it may appear to a GAN that a cat could have 3 legs. Thus, GANs producing cats with less than 4 legs could be justified. 

That is precisely why we have chosen non-overlapping squares for our experiment. Since there are no occlusions, the GAN should learn that in every image, there are supposed to be exactly 3 squares. Even in real world datasets, it may not be possible to collect a dataset which has cats whose legs have the same shape. However, it is possible to collect a dataset with cats having four legs in all images (no occlusions). 

Interestingly, both models seem to learn that there is only one square in the 1-4 dataset i.e. when trained on the Squares 1-4 dataset, most samples seem to include just 1 square of size which visually looks close to 4x4. Admittedly, this visual test of similarity may not be enough to ascertain whether the shape and size of the square corresponds exactly to a 4x4 square. However, during training with images of 1 square, we observed that if we increase the size of the square i.e. instead of using 4x4 use 16x16 (Squares 1-16), the likelihood of having multiple squares in the image reduces. 


We also observe that all squares generated seem to be axis aligned. Therefore, GANs seem to have no problem learning orientation but cannot seem to enforce counting constraints.

We also investigated whether a GAN can leverage the fact that it has learnt what a square looks like on Squares 1-4. In this experiment, we transferred the weights from that trained model and fine-tuned it on the Squares 3-4 dataset. We did not observe a noticeable improvement in image quality. If anything, image quality tends to get worse.

Samples from models trained on CT2 and additional samples from our models are included in the Appendix. We observed that the GANs did not sufficiently reproduce circles and triangles when trained on CT2. Since CT2 contains overlaps and multiple types of polygons, we consider this to be a more challenging dataset to model.

\section{Conclusion}

In this paper, we present the phenomenon of GANs being unable to count. We support this hypothesis with experiments on synthetic datasets where the count of similar objects in a scene is kept constant while their location is varied. We find that in their current form GANs are unable to learn semantic constraints even in the absence of noise introduced by natural image datasets. We also emphasize the fine line between generalization and good learning outcomes in GANs. 
Additionally, we conduct experiments on non-image data where we conclude that GANs tend to have difficulty learning discontinuous distributions which might necessitate the usage of mixtures of generators. A thorough evaluation of such an approach is left as future work. 

\bibliography{example_paper}
\bibliographystyle{icml2019}

\appendix
\section{Data Generation}

Here we describe the algorithms used for dataset creation.

\subsection{Point Datasets}

We used scikit-learn's \cite{scikit-learn} sklearn.datasets module for dataset creation. Each dataset has 5000 examples and consists of points in $R^2$ or $R^3$ for easier visualization. 

\begin{table}[h!]
  \begin{center}
    \begin{tabular}{c|c|c} 
      \textbf{Dataset} & \textbf{Function used} & \textbf{Dimension}\\
      \hline
      Circles & make\_circles & 2\\
      Mix of Gaussians & make\_blobs & 2\\
      S Curve & make\_s\_curve & 3\\
      Swiss Roll & make\_swiss\_roll & 3\\
    \end{tabular}
    \caption{Point Dataset summary}
    \label{table1}
  \end{center}
\end{table}

Each dataset has associated noise parameters corresponding to the noise parameters in the scikit-learn API. We experiment with varying noise but we find that it does not affect learning outcomes too much.

\subsection{Image Datasets}

The Squares datasets consist of non overlapping squares to test whether GANs can learn to count and to avoid introducing noise in the learning stage. To create images, we sample 3 random points in the image, which will serve as the top left point of the square. Then, we check whether squares drawn at these point overlap or not. If they do, then we again sample 3 new points otherwise we draw the squares to the image and add it to the dataset.

For the CT2 dataset, we use OpenCV's \cite{opencv_library} draw.ellipse() and draw.polygon() functions to draw figures at random point in the image. Note circles and traingles can overlap in this dataset, therefore, we consider this to be a more challenging dataset for a GAN to model.

\section{Architectural Details}

The architectures used in our experiments are given in the Tables \ref{table2}, \ref{table3}, \ref{table4} and \ref{table5}. We chose our architectures in accordance with common guidelines in GAN architectures \cite{DBLP:journals/corr/RadfordMC15, Gulrajani:2017:ITW:3295222.3295327, chintala}. Minor changes in the architecture (e.g changing ReLUs to Leaky ReLUs in the generator, adding or removing a few layers) did not seem to matter in terms of image quality in our experiments.

The size of the latent space for the image datasets was chosen to be 100 and for point datasets, it was chosen to be 2 for points in $R^2$ and 3 for points in $R^3$.

\begin{table}[h!]
  \begin{center}
    \begin{tabular}{c|c} 
      \textbf{Generator} & \textbf{Discriminator} \\
      \hline
      Dense & Dense\\
      Leaky ReLU & Leaky ReLU\\
      Dense & Dense\\
      Leaky ReLU & Leaky ReLU\\
      Dense & Dense\\
      Leaky ReLU & Leaky ReLU\\
      Dense & Dense\\
      Tanh & Sigmoid\\
    \end{tabular}
    \caption{Vanilla GAN Architecture for Points}
    \label{table2}
  \end{center}
\end{table}

\begin{table}[h!]
  \begin{center}
    \begin{tabular}{c|c} 
      \textbf{Generator} & \textbf{Discriminator} \\
      \hline
      Dense & Dense\\
      Leaky ReLU & Leaky ReLU\\
      Dense & Dense\\
      Leaky ReLU & Leaky ReLU\\
      Dense & Dense\\
      Leaky ReLU & Leaky ReLU\\
      Dense & Dense\\
      Tanh & -\\
    \end{tabular}
    \caption{WGAN-GP Architecture for Points}
    \label{table3}
  \end{center}
\end{table}

\begin{table}[h!]
  \begin{center}
    \begin{tabular}{c|c} 
      \textbf{Generator} & \textbf{Discriminator} \\
      \hline
      Dense & Conv2D\\
      Leaky ReLU & Leaky ReLU\\
      Reshape & Dropout\\
      Transposed Conv2D & Conv2D\\
      Batch Norm & Batch Norm\\
      Leaky ReLU & Leaky ReLU\\
      Transposed Conv2D & Dropout\\
      Batch Norm & Conv2D\\
      Leaky ReLU & Batch Norm\\
      Transposed Conv2D & Leaky ReLU\\
      Tanh & Dropout\\
      - & Conv2D\\
      - & Batch Norm\\
      - & Leaky ReLU\\
      - & Dropout\\
      - & Sigmoid\\
    \end{tabular}
    \caption{DCGAN Architecture for Images}
    \label{table4}
  \end{center}
\end{table}

\begin{table}[h!]
  \begin{center}
    \begin{tabular}{c|c} 
      \textbf{Generator} & \textbf{Critic} \\
      \hline
      Dense & Conv2D\\
      ReLU & Leaky ReLU\\
      Reshape & Dropout\\
      Transposed Conv2D & Conv2D\\
      Batch Norm & Batch Norm\\
      ReLU & Leaky ReLU\\
      Transposed Conv2D & Dropout\\
      Batch Norm & Conv2D\\
      Leaky ReLU & Batch Norm\\
      Transposed Conv2D & Leaky ReLU\\
      Tanh & Dropout\\
      - & Conv2D\\
      - & Batch Norm\\
      - & Leaky ReLU\\
      - & Dropout\\
      - & Dense\\
    \end{tabular}
    \caption{WGAN-GP Architecture for Images}
    \label{table5}
  \end{center}
\end{table}

\section{Additional Samples}

\begin{figure}
\centering
\begin{minipage}{.4\linewidth}
  \includegraphics[width=\linewidth]{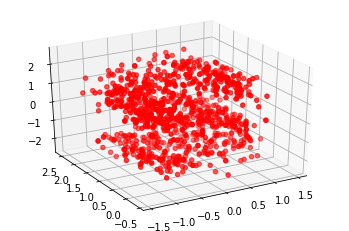}
  \caption{Original distribution of the S Curve (extra noise)}
  \label{img16}
\end{minipage}
\hspace{.05\linewidth}
\begin{minipage}{.4\linewidth}
  \includegraphics[width=\linewidth]{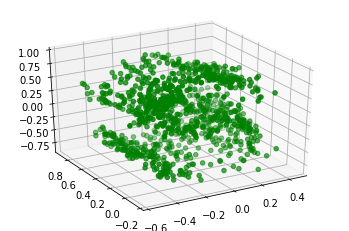}
  \caption{Learned Distribution after 150k steps (extra noise)}
  \label{img17}
\end{minipage}

\centering
\begin{minipage}{.4\linewidth}
  \includegraphics[width=\linewidth]{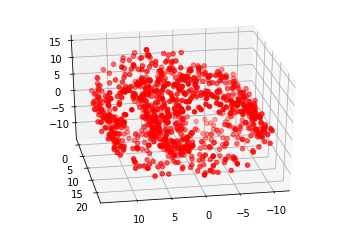}
  \caption{Original distribution of the Swiss Roll (extra noise)}
  \label{img18}
\end{minipage}
\hspace{.05\linewidth}
\begin{minipage}{.4\linewidth}
  \includegraphics[width=\linewidth]{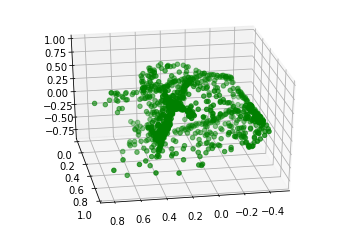}
  \caption{Learned Distribution after 150k steps (extra noise)}
  \label{img19}
\end{minipage}

\centering
    \includegraphics[width=\linewidth]{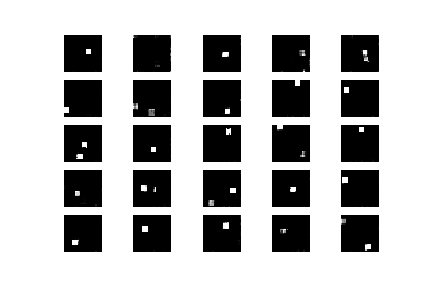}
    \vspace*{-1.2cm}
    \caption{Samples from model trained on Squares 1-4}
    \label{img20}
    
    \centering
    \includegraphics[width=\linewidth]{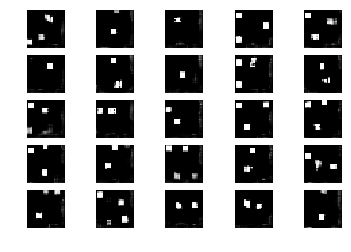}
    \vspace*{-1.2cm}
    \caption{Samples from models with transfer from Squares 1-4 to Squares 3-4}
    \label{img21}
\end{figure}

\begin{figure}
    \includegraphics[width=\linewidth]{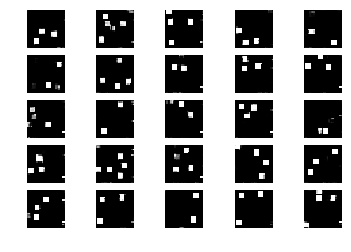}
    \vspace*{-1.2cm}
    \caption{Additional samples from DCGAN}
    \label{img22}
\end{figure}

\begin{figure}
    \includegraphics[width=\linewidth]{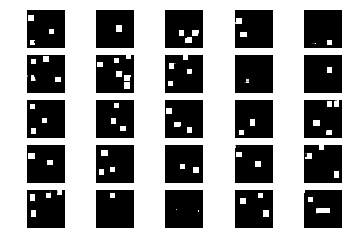}
    \vspace*{-1.2cm}
    \caption{Additional samples from WGAN-GP}
    \label{img23}
\end{figure}

\begin{figure}
    \includegraphics[width=\linewidth]{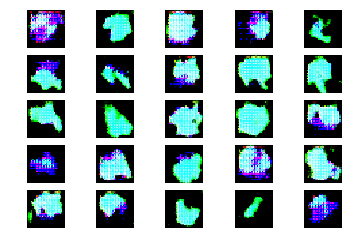}
    \vspace*{-1.2cm}
    \caption{Samples from model trained on CT-2 after 150k iters}
    \label{img24}
\end{figure}

Additional Samples from our models are shown in Figures \ref{img16}, \ref{img17}, \ref{img18}, \ref{img19}, \ref{img20}, \ref{img21}, \ref{img22}, \ref{img23}, \ref{img24}.

\end{document}